\documentclass{article}

\usepackage{microtype}
\usepackage{graphicx}
\usepackage{booktabs} 

\usepackage{helvet} 
\usepackage{courier}  
\usepackage[hyphens]{url}  
\usepackage{ulem}
\usepackage{subfig}
\usepackage{balance}
\usepackage{amssymb}
\usepackage{amsthm}
\usepackage{bbm}
\newtheorem*{remark}{Remark}

\usepackage[tbtags]{amsmath}
\newtheorem{definition}{Definition}

\DeclareMathOperator*{\argmin}{arg\,min}
\DeclareMathOperator*{\argmax}{arg\,max}

\usepackage{mathtools}
\DeclarePairedDelimiterX{\infdivx}[2]{(}{)}{%
  #1\;\delimsize\|\;#2%
}

\DeclarePairedDelimiter{\norm}{\lVert}{\rVert}
\newcommand{\Tau}{\mathcal{T}}

\newtheorem{theorem}{Theorem}[section]

\DeclareMathOperator{\rank}{rank}

\usepackage{multirow}

\usepackage{color, colortbl}
\definecolor{Gray}{gray}{0.9}
\usepackage[table]{xcolor}

\newcommand{\eat}[1]{}

\usepackage{hyperref}

\usepackage[accepted]{icml2020}

\icmltitlerunning{Progressive Graph Learning for Open-Set Domain Adaptation}

\begin{document}

\twocolumn[
\icmltitle{Progressive Graph Learning for Open-Set Domain Adaptation}



\icmlsetsymbol{equal}{*}

\begin{icmlauthorlist}
\icmlauthor{Yadan Luo}{equal,uq}
\icmlauthor{Zijian Wang}{equal,uq}
\icmlauthor{Zi Huang}{uq}
\icmlauthor{Mahsa Baktashmotlagh}{uq}
\end{icmlauthorlist}

\icmlaffiliation{uq}{School of Information Technology and Electrical Engineering, The University of Queensland, Australia}

\icmlcorrespondingauthor{Yadan Luo}{lyadanluol@gmail.com}

\icmlkeywords{Machine Learning, ICML}

\vskip 0.3in
]



\printAffiliationsAndNotice{\icmlEqualContribution} 

\begin{abstract}
Domain shift is a fundamental problem in visual recognition which typically arises when the source and target data follow different distributions. The existing domain adaptation approaches which tackle this problem work in the \textit{closed-set} setting with the assumption that the source and the target data share exactly the same classes of objects. In this paper, we tackle a more realistic problem of \textit{open-set} domain shift where the target data contains additional classes that are not present in the source data. More specifically, we introduce an end-to-end Progressive Graph Learning (PGL) framework where a graph neural network with episodic training is integrated to suppress underlying conditional shift and adversarial learning is adopted to close the gap between the source and target distributions. Compared to the existing open-set adaptation approaches, our approach guarantees to achieve a tighter upper bound of the target error. Extensive experiments on three standard open-set benchmarks evidence that our approach significantly outperforms the state-of-the-arts in open-set domain adaptation.
\end{abstract}

\section{Introduction}

While deep learning has made remarkable advances across a wide variety of machine-learning tasks and applications, it is commonly assumed that the training and test data are drawn from the same distribution. In practice, however, this assumption can be violated due to a number of factors, such as the change of lighting conditions, background, environment, or data modalities, which is referred to as the \textit{domain shift} problem. 

Unsupervised Domain Adaptation (UDA) approaches tackle the domain shift problem by aligning the training (source) and test (target) distributions, and can be roughly divided into statistical matching ~\cite{dip, manifold, distribution, DDC, JDA, DAN, DANN,ADDA,jingjing} or adversarial learning~\cite{DANN,ADDA,JAN,DRCN} methods. Theoretical analysis of UDA approaches has been widely studied~\cite{david1, discrepancy, bridgingtheory}, which provides rigorous error bounds on the target data.


Existing UDA algorithms are developed under the assumption that the source and target domains share an identical group of classes. Such a scenario typically refers to a \textit{closed-set} setting, which could be hardly guaranteed in real-world applications. Therefore, a more realistic Open-set adaptation setting has been introduced recently \cite{OSBP} which allows the target data to contain an additional ``unknown'' category, covering all irrelevant classes not present in the source domain.

The core idea of unsupervised open-set domain adaptation (OUDA)  approaches~\cite{ATI, mahsa,OSBP, STA, Attract-UTS} is to learn a classifier from a larger hypothesis space for both shared and unknown classes in the source and target domains. According to~\cite{david1, discrepancy}, the target error is bounded by the source risk, discrepancy distance across the domains, the shared error coming from the conditional shift~\cite{conditionalshift}, and the open-set risk. Open-set risk contributes the most to the error bound when a large percentage of data is unknown.

While promising, existing OUDA approaches~\cite{ATI,OSBP,mahsa,STA,Attract-UTS} lack an essential theoretical analysis of the aforementioned partial risks and the upper bound for the target risk, thus omitting potential solutions for improvement and leading to a biased solution. With the aim of minimising the aforementioned partial risks and achieving a tighter error bound for open-set adaptation, we combine the following four strategies in an end-to-end progressive learning framework:


    \begin{enumerate}
        \vspace{-1ex}
        \item To suppress the source risk, we decompose the original hypothesis space $\mathcal{H}$ into two subspaces $\mathcal{H}_1$ and $\mathcal{H}_2$, where $\mathcal{H}_1$ includes classifiers for the shared classes of the source and target domains and $\mathcal{H}_2$ is specific to classifying unknowns in the target domain. With a restricted size of the subspace $\mathcal{H}_1$, the possibility of misclassifying source data as unknowns will be reduced.
        
        \item To control the open-set risk, we adopt the progressive learning paradigm~\cite{curriculum}, where the target samples with low classification confidence are gradually rejected from the target domain and inserted as the pseudo-labeled unknown set in the source domain. This mechanism suppresses the potential negative transfer where the private representations across domains are falsely aligned. 
        
        \item We address conditional shift~\cite{conditionalshift} at the both sample- and manifold-level in a transductive setting. Specifically, we design an episodic training scheme and align conditional distributions across domains by gradually replacing the source data with the pseudo-labeled known data in each episode. We learn class-specific representations by aggregating the source and target features and passing episodes through deep graph neural networks.
        
        
        \item Our algorithm is seamlessly equipped with an adversarial domain discriminator, which effectively closes the gap between the source and target marginal distributions for the known categories.
       
    \end{enumerate}

We applied our method on three challenging open-set object recognition benchmarks, \textit{i.e.}, the \textit{Office-Home}, \textit{VisDA-17}, and \textit{Syn2Real-O}, to confirm its superiority to the existing state-of-the-art open-set domain adaptation approaches.







\section{Preliminaries}
    In this section, we introduce the notations, problem settings and the theoretical definitions and analysis for the tasks of closed-set and open-set unsupervised domain adaptation.
    
    \begin{definition}\textbf{Closed-set Unsupervised Domain Adaptation (UDA).} Let $\mathbb{P}^s$ and $\mathbb{Q}^t$ be the distributions of the source domain and the target domain, respectively. The corresponding label spaces for both domains are equal, \textit{i.e.}, $\mathcal{Y}_s = \mathcal{Y}_t$ = \{1, \ldots, C\}, where $C$ is the number of classes. The ultimate goal is to learn an optimal classifier $h\in \mathcal{H}$ for the target domain $h: \mathcal{X}_t\rightarrow\mathcal{Y}_t$,  based on the labeled source data and the unlabeled target data, where $\mathcal{H}$ is the hypothesis space of classifiers.
    \end{definition}

    \begin{definition}\textbf{Open-set Unsupervised Domain Adaptation (OUDA)~\cite{OSBP}.}
    Assume that we have the labeled source data $D_s = \{(x_{s_i}, y_{s_i})\}_{i=1}^{n_s}\sim \mathbb{P}^s$ and unlabeled target data $D_t = \{x_{t_j}\}_{j=1}^{n_t}\sim\mathbb{Q}^t_{X}$, where $\mathbb{P}^s$ being the joint probability distribution of the source domain, $\mathbb{Q}^t_X$ being the marginal distribution of the target domain, with $n_s$ and $n_t$ indicating the size of source and target dataset respectively. With training samples drawn i.i.d from both domains, the goal is to learn an optimal target classifier $h: \mathcal{X}_t\rightarrow \mathcal{Y}_t$. Here the target label space $\mathcal{Y}_t = \{\mathcal{Y}_s, unk\} = \{1, \ldots ,C+1\}$ includes the additional unknown class $C + 1$, which is not present in the source label space $\mathcal{Y}_s$.
    \end{definition}

    The source risk $R_s(h)$ and target risk $R_t(h)$ of a classifier $h\in\mathcal{H}$ with respect to the source distribution $\mathbb{P}^s$ and the target distribution $\mathbb{Q}^t$ are given by,
    \begin{align}
        R_s(h) &= \mathbb{E}_{(x, y)\sim\mathbb{P}^s}\mathcal{L}(h(x), y) = \sum_{i=1}^C\pi_{i}^s R_{s,i}(h),\nonumber\\
        R_t(h) &= \mathbb{E}_{(x, y)\sim\mathbb{Q}^t}\mathcal{L}(h(x), y) = \sum_{i=1}^{C+1}\pi_{i}^t R_{t,i}(h),
    \end{align}
    where $\pi_i^s = \mathbb{P}^s(y=i)$ and $\pi_i^t = \mathbb{Q}^t(y=i)$ are class-prior probabilities of the source and target distributions, respectively. The bounded loss function $\mathcal{L}: \mathcal{Y}_t\times \mathcal{Y}_t\rightarrow \mathbb{R}$ satisfies symmetry and triangle inequality. Particularly, the partial risk $R_{s,i}(h)$ and  $R_{t,i}(h)$ can be defined as,
    \begin{equation}
        \begin{split}
            R_{s,i}(h) &= \mathbb{E}_{x\sim\mathbb{P}^s(x|i)}\mathcal{L}(h(x), i),\\
            R_{t,i}(h) &= \mathbb{E}_{x\sim\mathbb{Q}^t(x|i)}\mathcal{L}(h(x), i).
        \end{split}
    \end{equation}
    \noindent Before introducing the generalization bound for open-set domain adaptation, it is crucial to define a discrepancy measure between the source and target domains:
    \begin{definition}\textbf{{Discrepancy Distance}~\cite{discrepancy}.} For any $h, h' \in \mathcal{H}$, the discrepancy between the distributions of the source and target domains can be formulated as:
    \begin{equation}
        \text{disc}(\mathbb{P}^s, \mathbb{Q}^t) = \sup_{h, h'\in\mathcal{H}}|\mathbb{E}_{\mathbb{P}^s}\mathcal{L}(h, h') - \mathbb{E}_{\mathbb{Q}^t}\mathcal{L}(h, h')|.
    \end{equation}
    
    \end{definition}
    
    \begin{theorem}\textbf{\textit{Open-set Domain Adaptation Upper Bound}}~\cite{open_theory}.\label{thm:OUDA}
    \label{open-set bound}
    Given the hypothesis space $\mathcal{H}$ with a mild condition that constant function $C + 1\in\mathcal{H}$, for $\forall h\in\mathcal{H}$, the expected error on target samples $R_t(h)$ is bounded as,
    \begin{equation}\label{eq:OUDA}
        \begin{split}
          \frac{R_t(h)}{1-\pi_{C+1}^t}&\leq R_s(h) + disc(\mathbb{Q}^t_{X|Y\leq C}, \mathbb{P}^s_X) + \lambda \\
              &+\underbrace{\frac{\pi_{C+1}^t}{1-\pi_{C+1}^t} R_{t, C+1}(h)}_{\text{open~set~risk}~\Delta_{o}},
        \end{split}
    \end{equation}
    \noindent where the shared error $\lambda =\min_{h\in\mathcal{H}}\frac{R_t^*(h)}{1-\pi_{C+1}^t} + R_s(h)$. The proof can be founded in the supplementary material.
    \end{theorem}
    \begin{remark}
    To compute the error upper bound for the closed-set unsupervised domain adaptation, Theorem \ref{thm:OUDA} can be reduced to:
    \begin{equation}
        \begin{split}
            R_t(h) &\leq R_s(h) +  disc(\mathbb{Q}^t_X, \mathbb{P}^s_X) + \lambda',\\
        \end{split}
    \end{equation}
    where $\pi_{C+1}^t = 0$ and $\lambda' = \min_{h\in\mathcal{H}}R_t(h)+R_s(h)$.
    \end{remark}

    According to Equation \eqref{eq:OUDA}, the target error is bounded by four terms, which opens four directions for improvement:
    \begin{itemize}
        \item Source risk $R_s(h)$. Assuming that source domain does not include any unknown samples, a part of the source risk can be avoided, which in turn minimizes the error upper bound. This direction is rarely investigated in the existing literature of open-set domain adaptation.
    
        \item Discrepancy distance $disc(\mathbb{Q}^t_{X|Y\leq C}, \mathbb{P}^s_X)$. Minimizing the discrepancy distance between the source and the target domains has been well investigated in recent years in statistics-based~\cite{MMD} or adversarial-based approaches~\cite{DANN}.
        
        \item Shared error $\lambda$ of the joint ideal hypothesis $h^*$. $\lambda$ tends to be large when the conditional shift encountered, where the class-wise conditional distributions are not aligned even with marginal distribution aligned.
        
        \item Open set risk $\Delta_{o}$. When a large percentage of data is unknown ($\pi_{C+1}^t\rightarrow 1$), this term contributes the most to the error bound. As shown in Equation \eqref{eq:OUDA}, it can be interpreted as the mis-classification rate for the unknown samples.
        
    \end{itemize}
    
     \begin{figure*}[t]
        \centering
        \includegraphics[width=0.98\linewidth]{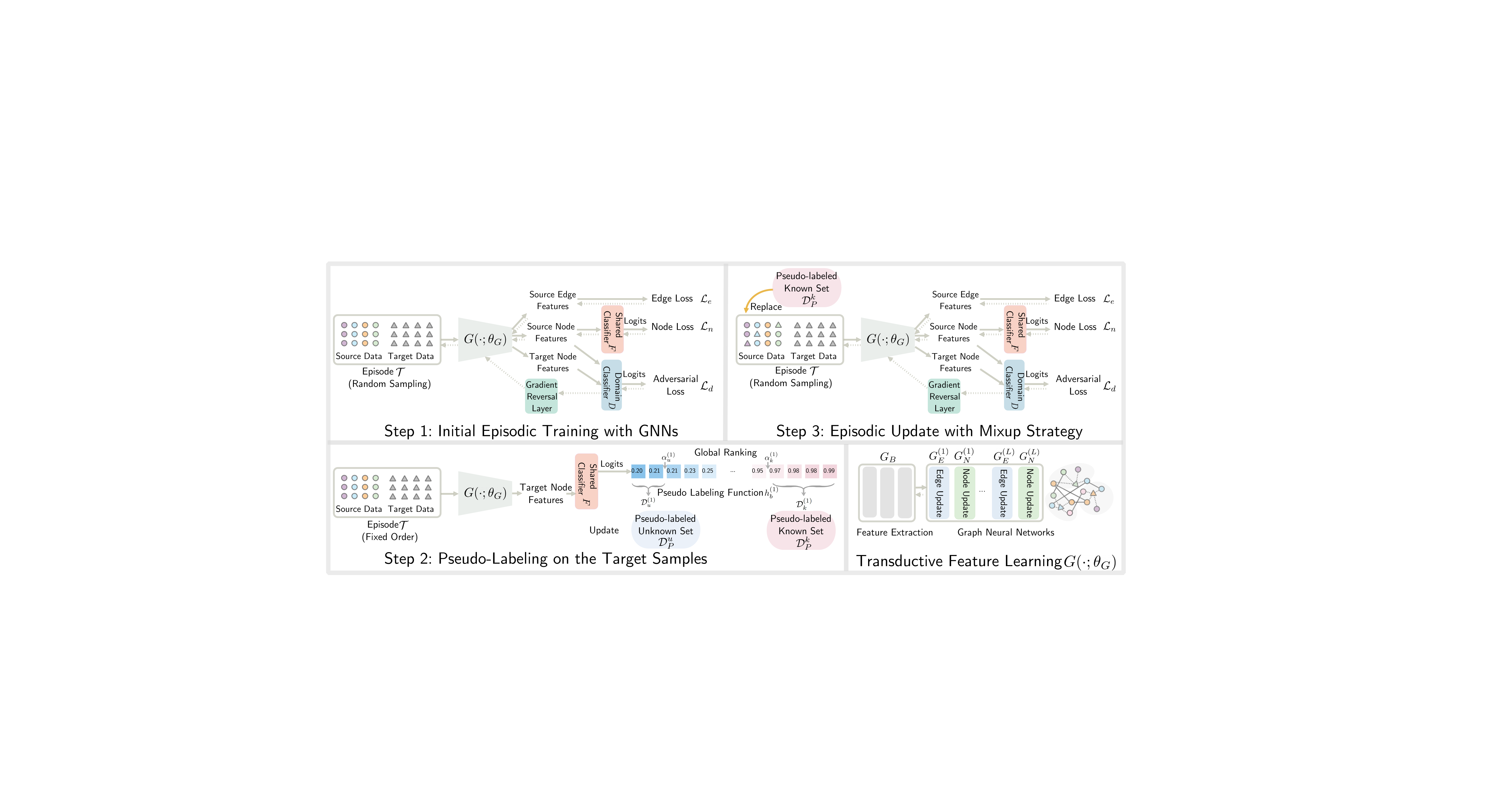}
        \caption{Proposed PGL framework. Circles indicate the source data and triangles are the target data. Different colors indicate different classes. By alternating between \textit{Steps 2} and \textit{3}, we progressively achieve the optimal classification model $F\circ G$ for the shared classes and pseudo-labeling function $h_b$ for rejecting the unknowns.}
        \label{fig:flowchart}
    \end{figure*}
    

    \section{Progressive Open-Set UDA}
        Aiming to minimise the four partial risks mentioned above, we reformulate the open-set unsupervised domain adaptation in a progressive way, and as such, we redefine the task at hand as follows.

       
    

    \begin{definition}\textbf{Progressive Open-Set Unsupervised Domain Adaptation (POUDA).}\label{def:PGL}
    Given the labeled source data $D_s = \{(x_{s_i}, y_{s_i})\}_{i=1}^{n_s}\sim \mathbb{P}^s$ and unlabeled target data $D_t = \{x_{t_j}\}_{j=1}^{n_t}\sim\mathbb{Q}^t_{X}$, the main goal is to learn an optimal target classifier $\tilde{h}\in \mathcal{H}_1\subset\mathcal{H}$ for the shared classes $Y_s = Y_t^* = \{1, \ldots, C\}$ and a pseudo-labeling function $h_b\in\mathcal{H}_2\subset\mathcal{H}$ for the unknown class $C+1$. 
    
    Assume the target set will be pseudo-labeled through $M$ steps, thereby the enlarging factor for each step is defined as $\alpha = \frac{1}{M}$. As long as the hypothesis $\tilde{h}$ and $h_b$ share the same feature extraction part, we can decompose the shared hypothesis $\tilde{h}$ into $\tilde{h}(x) = \argmax_{i\in Y_s}p(i|x)$ and define the pseudo-labeling function $h_b$ at the $m$-th step in line with $\tilde{h}$'s prediction:
    \begin{equation}\label{eq:rank}
        h_b^{(m)} = \begin{cases}
      C+1, & \text{if}~\rank(\max_{i\in Y_s}p(i|x))\leq\alpha_{u}^{(m)}, \\
      \tilde{y}, & \text{if}~\rank(\max_{i\in Y_s}p(i|x))\geq\alpha_{k}^{(m)},
    \end{cases}
    \end{equation} with $\alpha_u^{(m)} = \beta . \alpha . m . n_t$ and $\alpha_k^{(m)} = n_t - (1-\beta) . \alpha .  m . n_t$ being the index-based threshold to classify the unknown and known samples. The hyperparameter $\beta\in(0, 1)$ measures the openness of the given target set as the ratio of unknown samples. $\rank(\cdot)$ is a global ranking function which ranks predicted probabilities in ascending order and returns the sorted index list as an output. The pseudo-labeling function gives $\tilde{y} = \tilde{h}(x)$ for the possible known samples, and $C+1$ for the unknown ones. 
    \end{definition}
    
    In our case, the upper bound of expected target risk is formulated in the following theorem, 
    \begin{theorem}\textbf{\textit{POUDA Error Bound}.}
    \label{thm:POUDA}
    Given the hypothesis space $\mathcal{H}_1$, $\mathcal{H}_2\subset\mathcal{H}$, $\exists \alpha^*\in(0, 1)$, for $\forall \tilde{h}\in\mathcal{H}_1$ and $\forall h_b\in\mathcal{H}_2$, with a condition that the openness $\beta$ of the target set is fixed, the expected error $R_t(\tilde{h}, h_b)$ on the target samples is bounded as:

    \begin{equation}
        \begin{split}
          \frac{R_t(\tilde{h}, h_b)}{1-\pi_{C+1}^t}&\leq (1-\pi_{\alpha})(R_s(\tilde{h}) + disc(\mathbb{Q}^t_{X|Y\leq C}, \mathbb{P}^s_X)) + \lambda\\ &+\underbrace{\frac{\pi_{\alpha}\pi_{C+1}^t}{1-\pi_{C+1}^t}R_{t,C+1}(h_b)}_{\text{progressive~open~set~ risk}~\tilde{\Delta}_o} + const,
        \end{split}
    \end{equation}
    \noindent where the shared error $\lambda = \min_{\tilde{h}\in\mathcal{H}_1}\frac{R_t^*(\tilde{h})}{1-\pi_{C+1}^t} + (1-\pi_{\alpha})R_s(\tilde{h})$ and $\pi_{\alpha}$ indicates the probability that target samples being pseudo-labeled by $h_b$ (refer to the supplementary material for proof). 
    \end{theorem}
    
    \begin{remark}
     For $\tilde{h}\in\mathcal{H}_1\subset\mathcal{H}$ and $h_b\in\mathcal{H}_2\subset\mathcal{H}$, the following inequality holds, 
     \begin{equation}
        \begin{split}
        \sup_{\tilde{h}\in\mathcal{H}_1} R_s(\tilde{h}) &\leq \sup_{h\in\mathcal{H}} R_s(h),\\
        \sup_{h_b\in\mathcal{H}_2}\frac{\pi_{C+1}^t}{1-\pi_{C+1}^t} R_{t, C+1}(h_b) &\leq \sup_{h\in\mathcal{H}} \frac{\pi_{C+1}^t}{1-\pi_{C+1}^t} R_{t, C+1}(h).
        \end{split}
     \end{equation}
     We can observe that our progressive learning framework can achieve a tighter upper bound compared with conventional open-set domain adaptation framework. 
    \end{remark}


    \section{Methodology}
 
     In this section, we go through the details of the proposed Progressive Graph Learning (PGL) framework as illustrated in Figure \ref{fig:flowchart}. Our approach is mainly motivated by the two aspects of alleviating the shared error $\lambda$ and effectively controlling the progressive open-set risk $\tilde{\Delta}_o$ .

     \textbf{Minimizing the shared error $\lambda$.} \textit{Conditional shift}~\cite{conditionalshift} is the most significant obstacle for finding a joint ideal classifier for the source and target data, which arises when the class-conditional distributions of the input features substantially differ across the domains. That means, with unaligned distributions of the source distribution $\mathbb{P}_{X|Y}^s$ and target distribution $\mathbb{Q}_{X|Y\leq C}^t$, there is no guarantee to find an ideal shared classifier for both domains. Therefore, we address the conditional shift in a transductive setting from two perspectives: 
     \begin{itemize}
         \item \textit{Sample-level}: Motivated by~\cite{meta,meta1}, we adopt the episodic training scheme (Section \ref{sec:episodic}), and leverage the source samples from each class to ``support'' predictions on unlabeled data in each episode. With an enlarging labeled set through pseudo-labeling (Section \ref{sec:progressive}), we progressively update training episodes by replacing the source samples with pseudo-labeled target samples (Section \ref{sec:mixup}).

         \item \textit{manifold-level}: To regularize the class-specific manifold, we construct $L$-layer Graph Neural Networks (GNNs) on top of the backbone network $G_B(\cdot; \theta_{B})$ (\textit{e.g.}, ResNet), which consists of paired node update networks $G_N(\cdot; \theta_{N})$ and edge update networks $G_E(\cdot; \theta_{E})$. The source nodes and pseudo-labeled target nodes from the same class are densely connected, aggregating information though multiple layers.
     \end{itemize}

\begin{figure}
        \centering
        \includegraphics[width=0.9\linewidth]{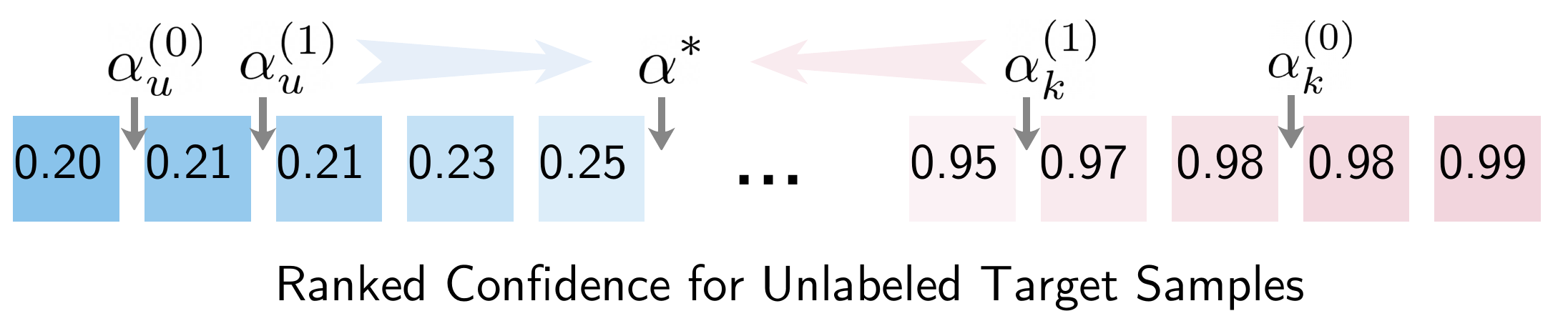}
        \caption{An illustration of the progressive learning to construct the pseudo-labeled target set. $\alpha^*$ indicates the ideal threshold for classifying known and unknown samples.}
        \label{fig:progressive}
    \end{figure}

     \textbf{Controlling progressive open-set risk $\tilde{\Delta}_o$.} As discussed in Section \ref{sec:progressive}, we iteratively squeeze the index-based thresholds $\alpha^{(m)}_u$ and $\alpha^{(m)}_k$ to approximate the optimal threshold $\alpha^*$ as illustrated in Figure \ref{fig:progressive}. Since the thresholds are mainly determined by the enlarging factor $\alpha$, we can always seek a proper value of $\alpha$ to alleviate the mis-classification error and the subsequent negative transfer. Our experimental results characterize the trade-off between computational complexity and performance improvement.


    \subsection{Initial Episodic Training with GNNs}\label{sec:episodic}
    Firstly, we denote the initial episodic formulation of a batch input as $\Tau^{(0)} = \{\Tau_s^{(0)}, \Tau_t^{(0)}\}=\{\tau_{s,i}^{(0)}, \tau_{t,i}^{(0)}\}_{i=1}^B$, with $B$ as the batch size. Each episode in the batch consists of two parts, \textit{i.e.}, the source episode $\tau_{s,i}^{(0)}  = \{(x_i, y)\}_{i=1}^C\sim \mathbb{P}^s_{X|Y}$ randomly sampled from each class $c\in Y_s$ and the target episode $\tau_{t,i}^{(0)} = \{x_j\}_{j=C+1}^{2C}\sim\mathbb{Q}^t_X$ randomly sampled from the target set. All instances in a mini-batch can form an undirected graph $\mathcal{G} = (\mathcal{V}, \mathcal{E})$. Each vertex $v_i\in\mathcal{V}$ is associated with a source or a target feature, and the edge $e_{ij}\in\mathcal{E}$ between nodes $v_i$ and $v_j$ measures the node affinity. The integrated GNNs are naturally able to perform a transductive inference taking advantage of labeled source data and unlabeled target data. The propagation rule for edge update and node update is elaborated in the following subsections.

    \subsubsection{Edge Update}
    The generic propagation rule for normalized edge features at the $l$-th layer can be defined as,
    \begin{align}\label{eq:adjacency}
            A_{ij}^{(l)} &= \sigma \big(G_E^{(l)}(\norm{v_i^{(l-1)} - v_j^{(l-1)}}; \theta_E^{(l)})\big),\nonumber\\ 
            \mathcal{E}^{(l)} &= D^{-\frac{1}{2}}(A^{(l)} + I) D^{-\frac{1}{2}},
    \end{align}
    with $\sigma$ being the sigmoid function, $D$ the degree matrix of $A^{(l)} + I$, $I$ the identity matrix, and $G_E^{(l)}(\cdot; \theta_E^{(l)})$ the non-linear network parameterized by $\theta_E$. 

    \subsubsection{Node Update}
    Similarly, the  propagation rule for node features at the $l$-layer is defined as,
    \begin{align}\label{eq:node}
        \hat{v}_i^{(l-1)}&= \sum_{j\in\mathcal{N}(i)}(v_i^{(l-1)} \mathcal{E}^{(l-1)}_{ij}),\nonumber\\
        v_i^{(l)} &= G_N^{(l)}([v_i^{(l-1)}; \hat{v}_i^{(l-1)}];\theta_N^{(l)}),
    \end{align}
    with $\mathcal{N}(i)$ being the neighbor set of the node $v_i$, $[\cdot;\cdot]$ the concatenation operation and $G_N^{(l)}(\cdot;\theta_N^{(l)})$ the network consisting of two convolutional layers, LeakyReLU activations and dropout layers. The node embedding is initialized with the extracted representations from the backbone embedding model, \textit{i.e.}, $v_i^{(0)}=G_B(x_i)$.

    \subsubsection{Joint Optimization}
    \textbf{Adaptive Learning.} We exploit adversarial loss to align the distributions of the source and target features extracted from the backbone network $G_B(\cdot;\theta_B)$. Specifically, a domain classifier $D(\cdot; \theta_D)$ is trained to discriminate between the features coming from the source or target domains, along with a generator $G_B$ to fool the discriminator $D$. The two-player minimax game shown in~Eq.\eqref{Eq:minmax} is expected to reach an equilibrium resulting in the domain invariant features:
    \begin{equation}
        \begin{split}
        \mathcal{L}_d = \mathbb{E}_{x\sim\Tau_s}\log [D(G_B(x))]+\mathbb{E}_{x\sim\Tau_t}\log [1 - D(G_B(x))].    
        \end{split}\nonumber
        \label{Eq:minmax}
    \end{equation}
    \textbf{Node Classification.} By decomposing the shared hypothesis $\tilde{h}$ into a feature learning module $G(\cdot,\theta_G)$ and a shared classifier $F(\cdot, \theta_F)$, we train the both networks to classify the source node embedding. To alleviate the inherent class imbalance issue, we adopt the focal loss to down-weigh the loss assigned to correctly-classified examples:
    \vspace{-2ex}
    \begin{equation}
        \mathcal{L}_n = -\mathbb{E}_{(x,y)\sim\Tau_s}\sum_{l=1}^{L}(1-F(G(x)^{(l)}))^\rho\log[F(G(x)^{(l)})],
        \nonumber
            \vspace{-2ex}
    \end{equation}
    with the hyperparameter $\rho = 2$ and $G(x)^{(l)}$ being the node embedding from the $l$-th node update layer. The total loss combines all losses from $L$ layers to improve the gradient flow in the lower layers.
    
    \textbf{Edge Classification.} Based on the given labels of the source data, we construct the ground-truth of edge map $\widehat{Y}$, where $\widehat{Y}_{ij} = 1$ if $x_i$ and $x_j$ belong to the same class, and $\widehat{Y}_{ij} = 0$ otherwise. The networks are trained by minimizing the following binary cross-entropy loss:
    \begin{equation}
        \mathcal{L}_e = -\mathbb{E}_{(x,y)\sim\Tau_s}\sum_{l=1}^{L} \widehat{Y} \log \mathcal{E}^{(l)} + (1-\widehat{Y})\log[1-\mathcal{E}^{(l)}].
        \nonumber
    \end{equation}
    
    \textbf{Final Objective Function.} Formally, our ultimate goal is to learn the optimal parameters for the proposed model,
    \begin{equation}\label{eq:optimization}
    \begin{split}
        (\theta^*_N, \theta^*_E, \theta^*_F, \theta^*_D) = \argmin \mathcal{L}_n + \mu\mathcal{L}_e + \gamma\mathcal{L}_d,\\
        (\theta^*_B) = \argmin \mathcal{L}_n + \mu\mathcal{L}_e - \gamma\mathcal{L}_d,
    \end{split}
    \end{equation}
    with $\mu$ and $\gamma$ the coefficients of the edge loss and adversarial loss, respectively.

    \subsection{Pseudo-Labeling in Progressive Paradigm}\label{sec:progressive}
    With the optimal model parameters obtained at the $m$-th step, we freeze the model and feed all the target samples in the forward direction, as shown in the \textit{Step 2} of Figure \ref{fig:flowchart}. Then, we rank the maximum likelihood $\max_{i\in Y_s}p(i|x)$ produced from the shared classifier $F(G(x)^{(L)})$ in an ascending order. Giving priority to the ``easier'' samples with relatively high/low confidence scores, we select $\alpha . m . n_t$ samples to enlarge the pseudo-labeled known set $\mathcal{D}_P^k$ and known set $\mathcal{D}_P^u$ (Refer to Eq.~\eqref{eq:rank}):
    \begin{equation}
        \begin{split}
            \mathcal{D}_P^k\leftarrow\mathcal{D}_k^{(0)}\cup\mathcal{D}_k^{(1)}\ldots\cup\mathcal{D}_k^{(m)},\\
            \mathcal{D}_P^u\leftarrow\mathcal{D}_u^{(0)}\cup\mathcal{D}_u^{(1)}\ldots\cup\mathcal{D}_u^{(m)},\\
            \mathcal{D}_k^{(m)} = \{(x_i, \tilde{y}_i)\}_{i=1}^{(1-\beta) . \alpha . m . n_t},\\
            \mathcal{D}_u^{(m)} = \{(x_j, C+1)\}_{j=1}^{\beta . \alpha . m. n_t}.
        \end{split}
    \end{equation}
    Note that $\mathcal{D}_k^{(m)}$ and $\mathcal{D}_u^{(m)}$ are newly annotated known set and unknown set, respectively and the pseudo-label is given by $\tilde{y}_i = \argmax_{y\in Y_s}p(y|x)$. To find a proper value of enlarging factor $\alpha$, we have two options: by aggressively setting a large value to $\alpha$, the progressive paradigm can be accomplished in fewer steps resulting in potentially noisy and unreliable pseudo-labeled candidates; on the contrary, choosing a small value of $\alpha$ can result in a steady increase of the model performance and the computational cost.
    
    \subsection{Episodic Update with Mix-up Strategy}\label{sec:mixup}
    We mix the source data with the samples from the updated pseudo-labeled known-set $\mathcal{D}^k_P$ at the $m$-th step, and construct new episodes $\Tau^{(m+1)}$ at the $(m+1)$-th step, as depicted in the \textit{Step 3} of Figure \ref{fig:flowchart}. In particular, We randomly replace the source samples with pseudo-labeled known data with a probability $\mathbb{P}^{(m)}_r=m\alpha$. Each episode in the new batch consists of three parts,
    \begin{equation}
    \begin{split}
        \tau_{s,i}^{(m+1)} &= \{(x_i, y_i)\}_{i=1}^{C\times(1-\mathbb{P}^{(m)}_r)}\sim \mathbb{P}^s(x|y),\\
        \widetilde{\tau}_{t,i}^{(m+1)} &= \{(x_k, \tilde{y}_k)\}_{k=1}^{C\times\mathbb{P}^{(m)}_r}\sim \mathbb{Q}^t(x|\tilde{y}), \\
        \tau_{t,i}^{(m+1)} &= \{x_j\}_{j=C+1}^{2C}\sim\mathbb{Q}^t_X,
    \end{split}
    \end{equation}
    with $\mathbb{Q}^t(x|\tilde{y})$ being the conditional distribution of the pseudo-labeled known set at the $m$-th step. Then, we update the model parameters according to Equation \eqref{eq:optimization} and repeat pseudo-labeling with the newly constructed episodes until convergence.

\section{Experiments}
In this section, we quantitatively compare our proposed model against various domain adaptation baselines on the \textit{Office-Home}, \textit{VisDA-17} and \textit{Syn2Real-O}  datasets. The baselines include four open-set domain adaptation methods of \textbf{ATI-$\lambda$} \cite{ATI}, \textbf{OSBP} \cite{OSBP}, \textbf{STA }\cite{STA}, \textbf{DAOD}~\cite{Attract-UTS}; two closed-set domain adaptation methods of \textbf{MMD}~\cite{MMD}, \textbf{DANN}~\cite{DANN}, and a basic \textbf{ResNet-50}~\cite{resnet} deep classification model. To be able to apply the closed-set baseline methods (ATI-$\lambda$, MMD, DANN, ResNet-50) in the open-set setting, we follow the previous baselines~\cite{STA,Attract-UTS} and reject unknown outliers from the target data using \textbf{OSVM}~\cite{OSVM} and \textbf{OSNN}~\cite{OSNN}.

\textbf{Evaluation Metrics:}
To evaluate the proposed method and the baselines, we utilize three widely used measures~\cite{OSBP, STA}, \textit{i.e.}, accuracy (\textbf{ALL}), normalized accuracy for all classes (\textbf{OS}) and normalized accuracy for the known classes only (\textbf{OS$^*$}):
\begin{equation}
\begin{split}
     &\text{ALL} = \frac{|x: x \in \mathcal{D}_t \land  h(x)=y|}{|(x, y): (x, y) \in \mathcal{D}_t|},\\
    &\text{OS} = \frac{1}{C+1}\sum_{i=1}^{C+1}\frac{|x: x \in \mathcal{D}^i_t \land h(x)=i|}{|x: x \in \mathcal{D}^i_t|},\\
    &\text{OS}^* = \frac{1}{C}\sum_{i=1}^{C}\frac{|x: x \in \mathcal{D}^i_t \land h(x)=i|}{|x: x \in \mathcal{D}^i_t|},\\
\end{split}
\end{equation}
with $\mathcal{D}_t^i$ being the set of target samples in the $i$-th class, and $h(\cdot)$ the classifier. In our case, we use the shared classifier $\tilde{h}$ for the known classes and pseudo-labeling function $h_b$ for the unknown one.

\textbf{Implementation Details:}
PyTorch implementation of our approach is avaibale in an annonymized repository\footnote{https://github.com/BUserName/PGL}. In our experiments, we employ ResNet-50 \cite{resnet} or VGGNet \cite{VGG} pre-trained on ImageNet as the backbone network. For VGGNet, we only fine-tune the parameters in FC layers. The networks are trained with the ADAM optimizer with a weight decay of $5\times10^{-5}$. The learning rate is initialized as $1\times10^{-4}$ and $1\times10^{-5}$ for the GNNs and the backbone module respectively, and then decayed by a factor of $0.5$ every $4$ epochs. The dropout rate is fixed to $0.2$ and the depth of GNN $L$ is set to $1$ for all experiments. The loss coefficients $\gamma$ and $\mu$ are empirically set to $0.4$ and $0.3$, respectively. The detailed experimental settings for the three datasets are summarized in Table \ref{tab:detail}, where $B$ being the batch size, $\alpha$ the enlarging factor, $\beta$ the hyperparameter for balancing the openness. The image feature extracted by the fc7 layer of VGGNet backbone is a 4096-D vector, and the deep feature extracted from the ResNet-50 is a 2048-D vector.
\begin{table}[t]
	\begin{center}
		\caption{The detailed settings for the three datasets used in our experiments.}\label{tab:detail}
		\vspace{1ex}
		\resizebox{0.48\textwidth}{!}{
		\begin{tabular}{l ccccc}
			\toprule
			Dataset &$B$ &Node Features  &Edge Features  &$\alpha$ &$\beta$\\
			\midrule
			\textit{Office-Home} &2 &512-D &512-D &0.05 & 0.6\\
			\midrule
			\textit{VisDA-17} &8 &1,024-D &1,024-D &0.05 &0.85\\
			\midrule
			\textit{Syn2Real-O} &6 &1,024-D &1,024-D &0.05 & 0.9\\
			\bottomrule
		\end{tabular}}
	\end{center}
			\vspace{-2ex}
\end{table}

\begin{table*}[!t]
	\begin{center}
		\caption{Recognition accuracies (\%) on 12 pairs of source/target domains from \textit{Office-Home} benchmark using ResNet-50 as the backbone. \textbf{Ar}: Art, \textbf{Cp}: Clipart, \textbf{Pr}: Product, \textbf{Rw}: Real-World. $^*$ indicates our re-implementation with the officially released code.}\label{tab:home} \vspace{1ex}
		\resizebox{1\textwidth}{!}{
		\begin{tabular}{lcccccccccccccccccccccccccc}
			\toprule
			\multirow{2}{*}{Method}&
			\multicolumn{2}{c}{\textbf{Ar$\to$Cl}}&
			\multicolumn{2}{c}{\textbf{Ar$\to$Pr}}&
			\multicolumn{2}{c}{\textbf{Ar$\to$Rw}}&
			\multicolumn{2}{c}{\textbf{Cl$\to$Rw}}&
			\multicolumn{2}{c}{\textbf{Cl$\to$Pr}}&
			\multicolumn{2}{c}{\textbf{Cl$\to$Ar}}&
			\multicolumn{2}{c}{\textbf{Pr$\to$Ar}}&
			\multicolumn{2}{c}{\textbf{Pr$\to$Cl}}&
			\multicolumn{2}{c}{\textbf{Pr$\to$Rw}}&
			\multicolumn{2}{c}{\textbf{Rw$\to$Ar}}&
			\multicolumn{2}{c}{\textbf{Rw$\to$Cl}}&
			\multicolumn{2}{c}{\textbf{Rw$\to$Pr}}&
			\multicolumn{2}{c}{\textbf{Avg.}}\\
			\cmidrule(l){2-3}
			\cmidrule(l){4-5}
			\cmidrule(l){6-7}
			\cmidrule(l){8-9}
			\cmidrule(l){10-11}
			\cmidrule(l){12-13}
			\cmidrule(l){14-15}
			\cmidrule(l){16-17}
			\cmidrule(l){18-19}
			\cmidrule(l){20-21}
			\cmidrule(l){22-23}
			\cmidrule(l){24-25}
			\cmidrule(l){26-27}
			& OS & OS$^*$& OS & OS$^*$& OS & OS$^*$& OS & OS$^*$& OS & OS$^*$& OS & OS$^*$& OS & OS$^*$& OS & OS$^*$& OS & OS$^*$& OS & OS$^*$& OS & OS$^*$& OS & OS$^*$& OS & OS$^*$\\
			\midrule
			ResNet+OSNN &33.7&32.1 &40.6 &39.4 &57.0 &56.6 &47.7 &46.9 &40.3 &39.1 &34.0 &32.3& 39.7 &38.5 &36.3&35.0 &59.7&59.6 & 52.1 & 51.4 &39.2 & 38.0 &59.2&59.2 & 45.0 & 44.0 \\
			ResNet+OSVM &37.5&38.7 &42.2 &42.6 &49.2 &51.4 &53.8 &55.5 &48.5 &50.0 & 39.2 &40.3 &53.4 &55.1 &43.5 &44.8 &70.6 &72.9 &65.6 &67.4 &49.5 & 50.8 &72.7 &75.1 &52.1 &53.7\\
			DANN+OSVM &52.3 &52.1 &71.3 &72.4 &82.3 &83.8 &73.2 &74.5 &62.8 &64.1 &61.4 &62.3 &63.5 &64.5 &46.0 &46.3 &77.2 &78.3 &70.5 &71.3 &55.5 &56.2 &79.1 &80.7 &66.2 &67.2\\
			ATI-$\lambda$+OSNN &53.1 &54.2 &68.6 &70.4 &77.3 &78.1 &74.3 &75.3 &66.7 &68.3 &57.8 &59.1 &61.2 &62.6 &53.9 &54.1 &79.9 &81.1 &70.0 &70.8 &55.2 &55.4 &78.3 &79.4 &66.4 &67.4\\
			\midrule
			OSBP &56.1 &57.2& 75.8 & 77.8 &83.0 &85.4 &75.5 &77.2 &69.2 &71.3 &64.6 &65.9 &64.6 &65.3 &48.3 &48.7 &79.5 &81.6 &72.1 &73.5 &54.3 &55.3 &80.2 &81.9 &68.6  &70.1 \\
			STA &58.1 &- & 71.6 &- &85.0 &- &75.8 &- &69.3 &- &63.4 &- &65.2 &- &53.1 &- &80.8 &- &74.9 &- &54.4 &- &81.9 &- &69.5&-\\
			STA$^*$ &46.6&45.9&67.0&67.2&76.2&76.6&64.9&65.2&57.7&57.6&50.2&49.3&49.5&48.4&42.9&40.8&76.6&77.3&68.7&68.6&46.0&45.4&73.9&74.5&60.0&59.8\\
			DAOD & 56.1 & 55.5 &69.1 & 69.2 &78.7&79.3 & 77.3& 78.2& 69.6& 70.2 &62.6 &62.9 &66.8 &67.7 &\textbf{59.7} & \textbf{60.3}& \textbf{83.3} &\textbf{85.0} & 72.3 &73.2 & 59.9 & 60.4 & 81.8 & 82.8 &69.8 &70.4\\
			\midrule
			\midrule
			\textbf{PGL}&\textbf{61.6} &\textbf{63.3} &\textbf{77.1} &\textbf{78.9} &\textbf{85.9} &\textbf{87.7} &\textbf{82.8} &\textbf{85.9} &\textbf{72.0} &\textbf{73.9} &\textbf{68.8} &\textbf{70.2} &\textbf{72.2} &\textbf{73.7} &58.4 & 59.2 &82.6 &84.8 &\textbf{78.6} &\textbf{81.5} &\textbf{65.0} &\textbf{68.8} &\textbf{83.0} &\textbf{84.8} &\textbf{74.0} &\textbf{76.1}\\
			\bottomrule
		\end{tabular}}
		\vspace{-2ex}
	\end{center}
\end{table*}

\begin{table*}[t]
	\begin{center}
		\caption{Recognition accuracies (\%) for open-set domain adaptation experiments on the \textit{Syn2Real-O} (ResNet-50). }\label{tab:Syn2Real-O} \vspace{1ex}
		\resizebox{0.92\textwidth}{!}{
		\begin{tabular}{l c c c c c c c c c c c c c c c c}
			\toprule
			\multirow{1}{*}{Method}
			&Aer&
			Bic&
			Bus&
			Car&
			Hor&
			Kni&
			Mot&
			Per&
			Pla&
			Ska&
			Tra&
			Tru&
			UNK&
			OS&
			OS$^*$\\
			\midrule
			ResNet~\cite{resnet}+OSVM &29.7 &39.2 &49.9 &54.0 &76.8 &22.2 &71.2&32.6&75.1&21.5&65.2&0.6&45.2&44.9&44.8\\
			DANN~\cite{DANN}+OSVM & 50.8&44.1&19.0&58.5&76.8 &26.6&68.7 &\textbf{50.5}&82.4&21.1&69.7&1.1&33.6&46.3&47.4\\
			\midrule
			OSBP~\cite{OSBP} &75.5 &67.7 &68.4 &\textbf{66.2}&71.4&0.0&86.0&3.2&39.4&23.2&68.1&3.7&\textbf{79.3}&50.1&47.7\\
			STA~\cite{STA} & 64.1& \textbf{70.3}& 53.7& 59.4 & 80.8& 20.8 & 90.0& 12.5& 63.2& 30.2& 78.2& 2.7&59.1&52.7&52.2\\
			\midrule
			\midrule
			\textbf{PGL} &\textbf{81.5} &68.3 &\textbf{74.2} & 60.6 &\textbf{91.9} &\textbf{45.4} &\textbf{92.2} &41.0 &\textbf{87.9} &\textbf{67.5} &\textbf{79.2} &\textbf{6.4} &49.6 &\textbf{65.5} &\textbf{66.8}\\
    		\bottomrule
		\end{tabular}
		}
	\end{center}
\end{table*}

\subsection{Datasets}

\textbf{Office-Home}~\cite{officehome} is a challenging domain adaptation benchmark, which comprises 15,500 images from 65 categories of everyday objects. The dataset consists of 4 domains: Art (\textbf{Ar}), Clipart (\textbf{Cp}), Product (\textbf{Pr}), and Real-World (\textbf{Rw}). Following the same splits used in ~\cite{STA}, we select the first 25 classes in alphabetical order as the known classes, and group the rest of the classes as the unknown. 

\textbf{VisDA-17}~\cite{visda2017} is a cross-domain dataset with 12 categories in two distinct domains. The \textbf{Synthetic} domain consists of 152,397 synthetic images generated by 3D rendering and the \textbf{Real} domain contains 55,388 real-world images from MSCOCO~\cite{MSCOCO} dataset. Following the same protocol used in ~\cite{OSBP, STA}, we construct the known set with 6 categories and group the remaining 6 categories as the unknown set. 

\textbf{Syn2Real-O}~\cite{visda18} is the most challenging synthetic-to-real testbed, which is constructed from the \textit{VisDA-17}. The Syn2Real-O dataset significantly increases the openness to 0.9 by introducing additional unknown samples in the target domain. According to the official setting, the \textbf{Synthetic} source domain contains training data from the \textit{VisDA-17} as the known set, and the target domain \textbf{Real} includes the test data from the \textit{VisDA-17} (known set) plus 50k images from irrelevant categories of MSCOCO dataset (unknown set).





\subsection{Results and Analysis}
As reported in Table~\ref{tab:home}, Table~\ref{tab:Syn2Real-O}, and Table~\ref{tab:visda17}, we clearly observe that our method \textbf{PGL} consistently outperforms the state-of-the-art results, improving mean accuracy (OS$^*$) by $8.1\%$, $28.0\%$ and $29.6\%$ on the benchmark datasets of \textit{Office-Home}, \textit{Syn2Real-O} and \textit{VisDA-17} datasets respectively. Note that our proposed approach provides significant performance gains for the more challenging datasets of \textit{Syn2Real-O} and \textit{VisDA-17} which require knowledge transfer across different modalities. This phenomenon can be also observed in the transfer sub-tasks with a large domain shift \textit{e.g.}, \textbf{Rw}$\to$\textbf{Cl} and \textbf{Pr}$\to$\textbf{Ar} in \textit{Office-Home}, which demonstrates the strong adaptation ability of the proposed framework. To study the validity of the progressive paradigm and early stopping strategy, we provide detailed graphs of our test performance per training step (OS, OS$^*$ and ALL scores) in the supplementary material.

\begin{table}[t]
	\begin{center}
	\vspace{-2ex}
		\caption{Performance comparisons on the \textit{VisDA-17} (VGGNet).}\label{tab:visda17} \vspace{1ex}
		\resizebox{0.47\textwidth}{!}{
		\begin{tabular}{l c c c c c c c c c }
			\toprule
			Method
			&Bic &Bus &Car &Mot &Tra &Tru &UNK &OS &OS$^*$\\
			\midrule
			MMD+OSVM &39.0 &50.1 &64.2 &79.9 &86.6 &16.3 &44.8 &54.4 &56.0 \\
			DANN+OSVM &31.8 &56.6 &71.7 &77.4 &87.0 &22.3 &41.9 &55.5 &57.8\\
			ATI-$\lambda$+OSVM &46.2 &57.5& 56.9& 79.1& 81.6& 32.7 &65.0 & 59.9 & 59.0\\
			OSBP & 51.1 & 67.1 & 42.8 & 84.2& 81.8 & 28.0 &\textbf{85.1} & 62.9& 59.2\\
			STA & 52.4 & 69.6 & 59.9 & 87.8 & 86.5 & 27.2 &84.1 & 66.8 & 63.9\\
			\midrule
			\midrule
			\textbf{PGL} &\textbf{93.5} &\textbf{93.8} &\textbf{75.7} &\textbf{98.8} &\textbf{96.2} &\textbf{38.5}&68.6 &\textbf{80.7} &\textbf{82.8}\\
			\bottomrule
		\end{tabular}}
	\end{center}
\end{table}

\begin{table}
	\begin{center}
		\caption{Ablation performance on the \textit{Syn2Real-O} (ResNet-50). ``w'' indicates with and ``w/o'' indicates without. }\label{tab:abl} \vspace{1ex}
		\resizebox{0.4\textwidth}{!}{
		\begin{tabular}{l cc cc}
			\toprule
    		Model
			&UNK & ALL & OS& OS$^*$\\
			\midrule
			PGL w/o Progressive &43.6&44.8&54.4 & 55.3\\
			PGL w NLL & 48.6 &49.7 &56.9 &57.6\\
			PGL w/o GNNs & 49.2&50.3& 57.8& 58.5\\
			PGL w/o Mix-up &\textbf{49.8}&51.3&62.5&63.6\\
			\midrule
			\midrule
		    \textbf{PGL}  &49.6 &\textbf{51.5}&\textbf{65.5} &\textbf{66.8}\\
			\bottomrule
		\end{tabular}}
	\end{center}
	\vspace{-2ex}
\end{table}

\begin{table}
\vspace{-2ex}
	\begin{center}
		\caption{Performance comparisons \textit{w.r.t.} varying enlarge factor $\alpha$ on the \textit{VisDA-17} (ResNet-50).}\label{tab:EF} \vspace{1ex}
		\resizebox{0.42\textwidth}{!}{
		\begin{tabular}{c cc cc}
			\toprule
			\multirow{2}{*}{\textbf{Enlarging Factor}}&
			\multicolumn{2}{c}{~~~~\textbf{Syn2Real-O}~~~~}&
			\multicolumn{2}{c}{\textbf{Office-Home (Ar-Cl)}}\\
			\cmidrule(l){2-3}
			\cmidrule(l){4-5}
			&~OS~~& OS$^*$& ~~OS~~~~~& OS$^*$\\
			\midrule
			$\alpha = 0.20$ & 63.0 & 63.3 & 59.9 & 61.1\\
			$\alpha  = 0.10$ & 64.5 & 65.7 & 60.7 & 61.6\\
		    $\alpha  = 0.05$ & \textbf{65.6} &\textbf{ 66.5} & \textbf{61.8} & \textbf{63.1}\\
			\bottomrule
		\end{tabular}}
	\end{center}
\end{table}


\begin{figure*} 
	\centering
	\includegraphics[width=1\linewidth]{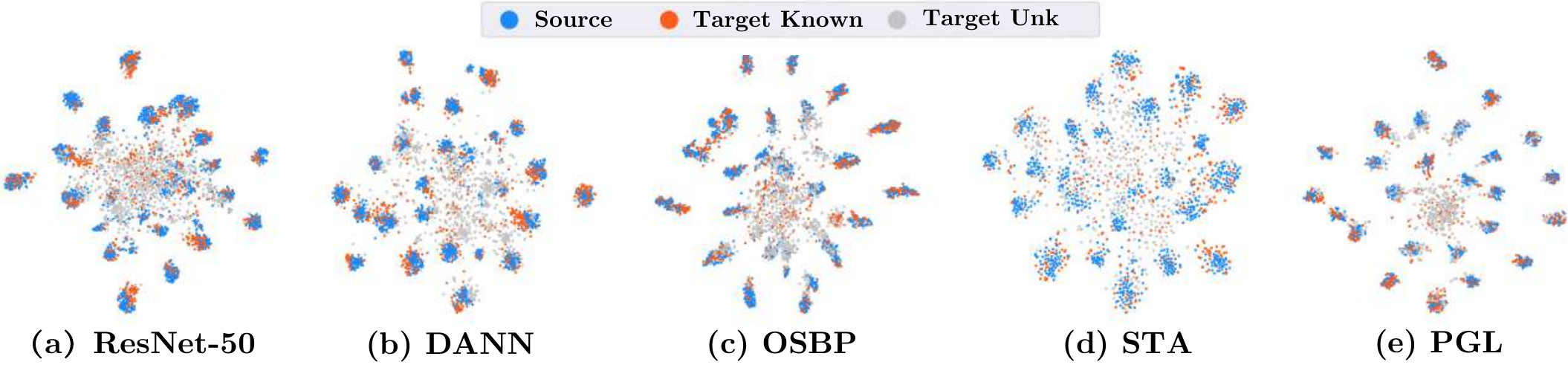}
	\vspace{-2ex}
    \centering
	\caption{The t-SNE visualization of feature distributions on the Rw$\to$Ar task (\textit{Office-Home}) with the ResNet-50 backbone.}\label{fig:tsne}
\end{figure*}

\begin{figure}[t]
	\centering
	\includegraphics[width=1\linewidth]{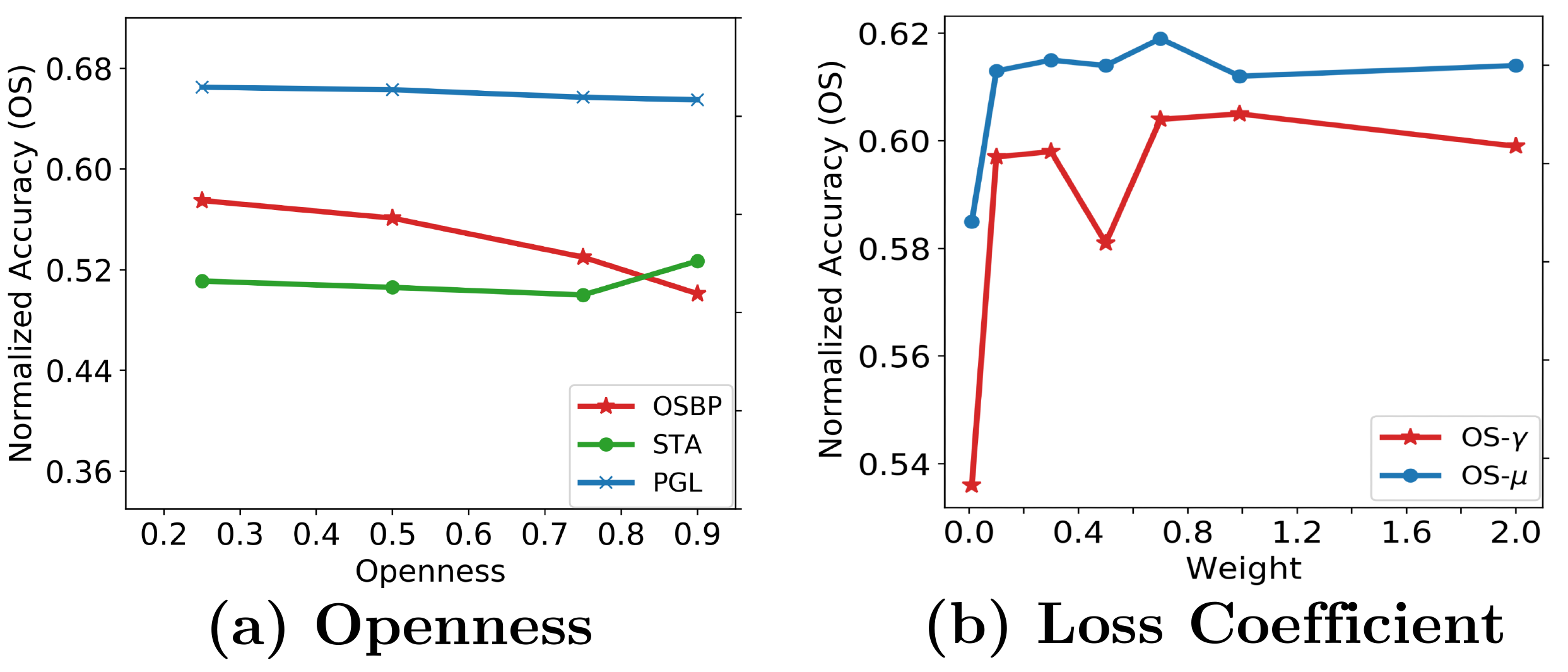}
	\caption{Performance Comparisons \textit{w.r.t.} varying (a) openness of the \textit{Syn2Real-o} (ResNet-50); (b) loss coefficients $\mu$ and $\gamma$ on the Ar$\to$Cl task (\textit{Office-Home}) with the ResNet-50 backbone.} \label{fig:analysis}
	\vspace{-1ex}
\end{figure}

\textbf{Ablation Study:}
To investigate the impact of the derived progressive paradigm, GNNs, node classification loss, and mix-up strategy, we compare four variants of the PGL model on the \textit{Syn2Real-O} dataset shown in Table \ref{tab:abl}. Except for \textbf{PGL w/o Progressive} that takes $\alpha = 1$ and $\beta=0.6$, all experiments are conducted under the default setting of hyperparameters. \textbf{PGL w/o Progressive} corresponds to the model directly trained with one step, followed by pseudo-labeling function for classifying the unknown samples. As shown in Table \ref{tab:abl}, without applying the progressive learning strategy, the OS result of \textbf{PGL w/o Progressive} significantly drops by 16.9\% because \textbf{PGL w/o Progressive} does not leverage the pseudo-labeled target samples leading to the failure in minimizing the shared error at the sample-level. In \textbf{PGL w NLL}, the focal loss of the node classification objective is replaced with the Negative log-likelihood (NLL) loss, resulting in OS performance dropping from 65.5\% to 56.9\%. Due to the absence of the focal loss re-weighting module, the model tends to assign more pseudo-labels to easy-to-classify samples, which consequently hinders effective graph learning in the episodic training process. In \textbf{PGL w/o GNNs}, we used ResNet-50 as the backbone for feature learning, which triggers 12.5\% OS performance drops comparing to the graph learning model. The inferior results reveal that the GNN module can learn the class-wise manifold, which mitigates the potential noise and permutation by aggregating the neighboring information. \textbf{PGL w/o Mix-up} refers to the model that constructs episodes without taking any pseudo-labeled target data. We observe that the OS performance of \textbf{PGL w/o Mix-up} is 4.6\% lower than the proposed model, confirming that replacing the source samples with pseudo-labeled target samples progressively can alleviate the side effect of conditional shift.

\textbf{Robustness Analysis to Varying Openness:}
To verify the robustness of the proposed PGL, we conduct experiments on the \textit{Syn2Real-O} with the openness varying in $\{0.25, 0.5, 0.75, 0.9\}$. The openness is defined as the ratio of unknown samples to all samples in the entire target set, which explicitly implies the level of challenge. The results of \textbf{OSBP}, \textbf{STA} and the proposed \textbf{PGL} are depicted in Figure~\ref{fig:analysis}(a). Note that \textbf{OSBP} and our \textbf{PGL} approach empirically sets a hyperparameter ($\beta$ in our case) to control the openness, while \textbf{STA} automatically generates the soft weight in adversarial way and inevitably results in performance fluctuation. We observe that \textbf{PGL} consistently outperforms the counterparts by a large margin, which confirms its resistance to the change in openness.

\begin{figure*}[t]
	\centering
	\includegraphics[width=0.97\linewidth]{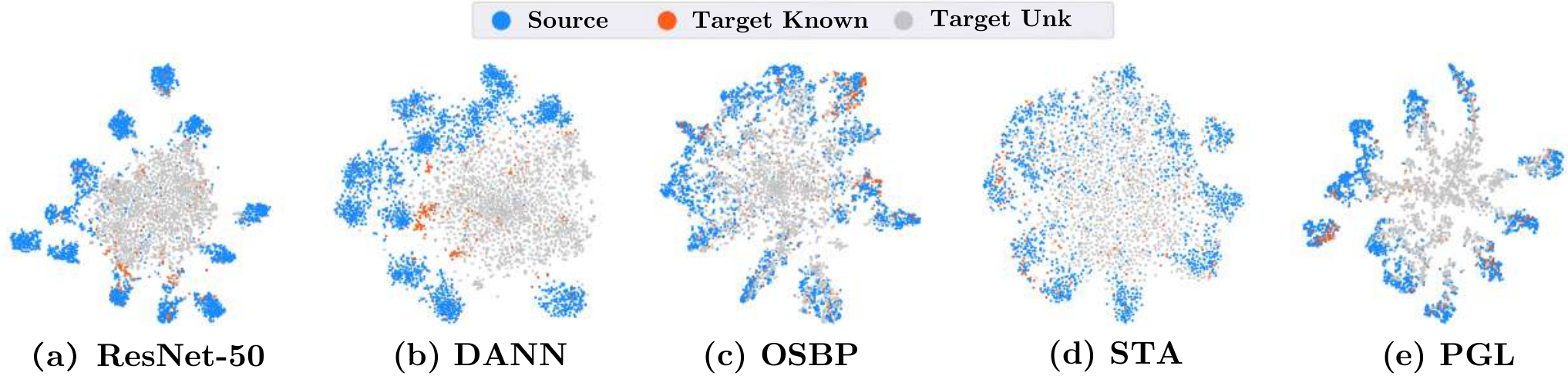}
	\caption{The t-SNE visualization for the source and target data in the \textit{Syn2Real-O} dataset.}\label{fig:tsne}
\end{figure*}

\textbf{Sensitivity to Loss Coefficients $\mu$ and $\gamma$:}
We show the sensitivity of our approach to varying the edge loss coefficient $\mu$ and adversarial loss coefficient $\gamma$ in Figure \ref{fig:analysis}(b). We vary the value of one loss coefficient from (0, 2] at each time, while fixing the other parameter to the default setting. Two observations can be drawn from Figure \ref{fig:analysis}(b): The OS score becomes stable when loss coefficients are within the interval of [0.7, 2]; When $\mu\rightarrow 0$, $\gamma\rightarrow 0$, the model performance drops by $4.6\%$ and $10.2\%$ respectively, which verifies the importance of the edge supervision and adversarial learning in our framework.

\begin{figure}[t]
	\centering
	{\includegraphics[width=.92\linewidth]{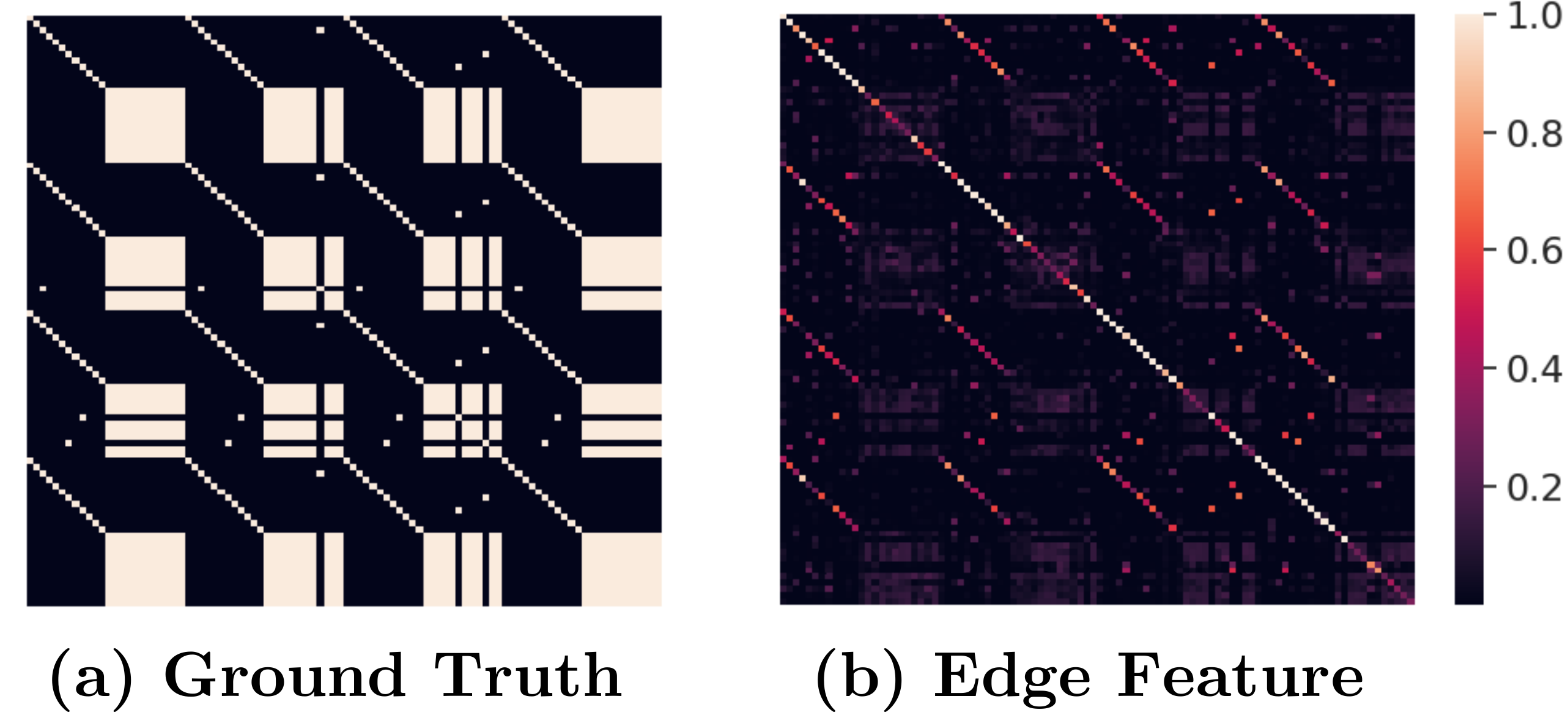}}
	\caption{Visualization of edge features on the \textit{Syn2Real-O}. \textit{Left}: the binary ground-truths label map. \textit{Right}: the learned edge map from the proposed edge update networks.  Best viewed in color.} \label{fig:edge}
\end{figure}

\textbf{Sensitivity to Enlarging Factor $\alpha$:}
We further study the effectiveness of the enlarging factor $\alpha$, which controls the enlarging speed of the pseudo-labeled set, shown in Table \ref{tab:EF}. We note that the proposed model with a smaller value of $\alpha$ consistently performs better on both the \textit{Syn2Real-O} and \textit{Office-Home} datasets. This testifies our theoretical findings that the progressive open-set risk $\tilde{\Delta}_o$ can be controlled by consecutively classifying unknown samples. With a sacrifice on the training time, this strategy also provides more reliable pseudo-labeled candidates for the shared classifier learning preventing the potential error accumulation in the next several steps.

\textbf{t-SNE Visualization.} To intuitively showcase the effectiveness of OUDA approaches, we extract features from the baseline models (\textbf{ResNet-50}, \textbf{DANN}, \textbf{OSBP}, \textbf{STA}) and our proposed model \textbf{PGL}
on the Rw$\to$Ar task (\textit{Office-Home}) with the ResNet-50 backbone. The feature distributions are visualized with t-SNE afterwards. As shown in Figure \ref{fig:tsne}, compared with \textbf{ResNet-50} and \textbf{DANN}, open-set domain adaptation methods generally have a better separation between the known (in blue and red) and unknown (in grey) categories. \textbf{STA} achieves a better alignment between the source and target distributions in comparison with \textbf{OSBP}, while the \textbf{PGL} can obtain a clearer class-wise classification boundary benefiting from our graph neural networks and the mix-up strategy. We conduct additional experiments on the challenging \textit{Syn2Real-O} dataset with a high openness. We randomly sample 200 episodes from the dataset including 2,400 source points and 2,400 target points and visualize the distributions of the learned representations from the compared baseline models and the proposed model in Figure \ref{fig:tsne}. Figure \ref{fig:tsne}(c)-(e) shows that the open-set domain adaptation methods are more robust to disturbance from unknowns compared with \textbf{ResNet-50} and \textbf{DANN}, as the source data (shown in blue) and the target data from the shared classes (shown in red) are aligned. A comparison between Figure \ref{fig:tsne}(d) and Figure \ref{fig:tsne}(e) reveals that the proposed graph learning and mix-up strategy have the ability to align class-specific conditional distributions across the domains, which means, the representations of the source and target data belonging to the same class are well mixed and less distinguishable.

\begin{figure}[!htb]
	\centering
	\subfloat[][Office-Home]
	{\includegraphics[width=0.5\linewidth]{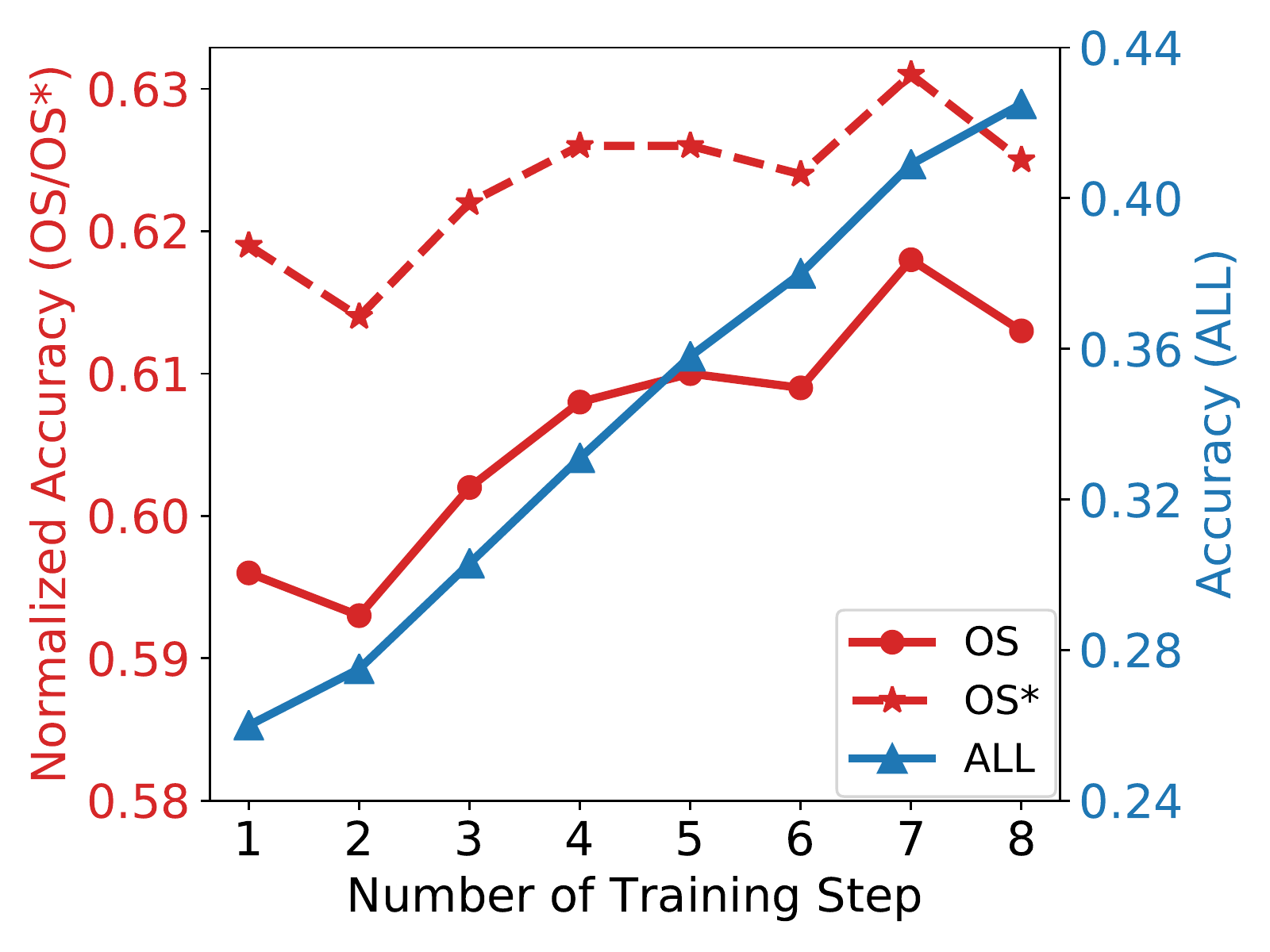}}
	\subfloat[][Syn2Real-O]
	{\includegraphics[width=0.5\linewidth]{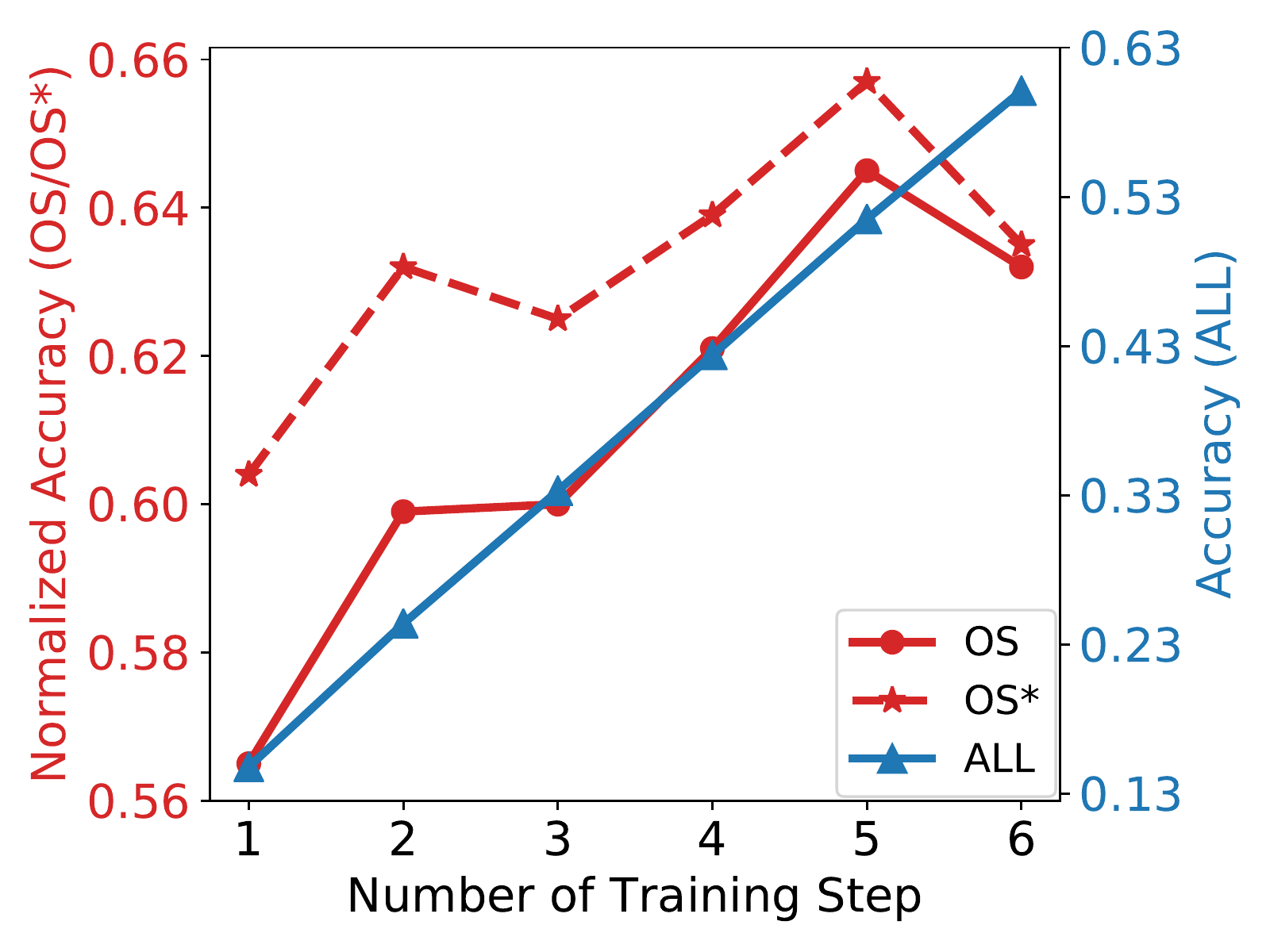}}
	\caption{Recognition accuracies of the proposed PGL method on the (a) Ar$\to$Cl task (\textit{Office-Home}) and (b) \textit{Syn2Real-O} datasets.} \label{fig:analysis}
\end{figure}
\textbf{Edgemap Visualization.} To further analyze the validity of the edge update networks, we extract the learned feature map  from the PGL with a single-layer GNN on the \textit{Syn2Real-O} dataset. As visualized in Figure \ref{fig:edge}(b), a large value of $\mathcal{E}_{ij}$ corresponds to a high degree of correlations between node $v_i$ and $v_j$, which resembles the pattern of the ground-truth edge label $\widehat{Y}$ as displayed in Figure \ref{fig:edge}(a).

\textbf{Quantitative Analysis over Training Steps.} Figure~\ref{fig:analysis} illustrates the recognition performance of PGL over training steps on  the Ar$\to$Cl task of the \textit{Office-Home} dataset and \textit{Syn2Real-O} dataset, respectively. Three evaluation metrics are used to testify performance, \textit{i.e.}, the overall accuracy \textbf{ALL}, and normalized accuracies  \textbf{OS} and \textbf{OS$^*$}. All metrics gain a performance over the first several steps as the pseudo-labeled target samples added in the source episodes can assist the classifier to make a more accurate prediction. Then, the normalized accuracy OS and OS$^*$ experience a downward because the enlarging pseudo-labeled set brings along noise and disturbance, which may degrade the model performance. In contrast, the accuracy ALL continuously increases as more unknown target samples are correctly classified, which occupy a large portion in the target domain. The results characterize a trade-off between normalized accuracy OS / OS$^*$ and the accuracy for unknowns. Considering the core value of domain adaptation is to correctly classify the classes of interest rather than irrelevant classes, we choose to stop the model updates at the training step 7 for the \textit{Office-Home} and the step 5 for the \textit{Syn2Real-O}.

\section{Conclusion}
We have addressed the open-set domain shift problem in both sample- and manifold-level by controlling the open-set risk. Experiments show that our proposed progressive graph learning framework performs consistently well on challenging object recognition benchmarks for open-set adaptation with significant domain discrepancy and conditional shifts. 

\section*{Acknowledgements}
This work was partially supported by ARC DP 190102353.

\balance
\newpage
\bibliography{main}
\bibliographystyle{icml2020}





\end{document}